\documentclass[letterpaper, 10 pt, journal, twoside]{IEEEtran}
\usepackage{amsmath,amsfonts}
\usepackage{stix}
\usepackage{algorithmic}
\usepackage{algorithm}
\usepackage{array}
\usepackage[caption=false,font=normalsize,labelfont=sf,textfont=sf]{subfig}
\usepackage{textcomp}
\usepackage{stfloats}
\usepackage{url}
\usepackage{verbatim}
\usepackage{graphicx}
\usepackage{cite}
\usepackage[table]{xcolor}

\usepackage{booktabs}
\usepackage{siunitx}
\usepackage{multirow}
\usepackage{hyperref}
\usepackage{cleveref}
\usepackage{graphicx} 
\renewcommand{\baselinestretch}{0.982}
\usepackage{caption}
\usepackage[font={scriptsize}]{caption}

\usepackage{float}

\DeclareMathOperator*{\argmin}{arg\,min}

\newcommand{\todo}[1]{\textbf{\color{red}[TODO: #1]}}

\definecolor{hermann}{rgb}{1.0,0.5,0.0}

\definecolor{Tianyi}{rgb}{0.08,0.6,0.33}
\newcommand{\Tia}[1]{\textcolor{Tianyi}{\emph{TianYi:~{#1}}}}
\definecolor{boyang}{rgb}{0.1,0.1,0.8}
\newcommand{\boy}[1]{\textcolor{boyang}{\emph{Boyang:~{#1}}}}

\definecolor{odomEdge}{HTML}{93CCEA}
\definecolor{camEdge}{HTML}{dd7e6b}
\definecolor{lcEdge}{HTML}{38761d}

\usepackage{soul} 
\usepackage[normalem]{ulem} 

\begin{document}
%
\title{Loop Closure from Two Views:\\Revisiting PGO for Scalable Trajectory Estimation through Monocular Priors}
%
%
%

\author{Tian~Yi~Lim$^{*\wedge1}$, Boyang~Sun$^{*2}$, Marc~Pollefeys$^{2,3}$, and Hermann~Blum$^{2,4}$%
\thanks{$^{*}$ equal contribution. $^{\wedge}$ Work done at $^{2}$}%
\thanks{$^{1}$ DSO National Laboratories, $^{2}$ ETH Zurich, $^{3}$ Microsoft, $^{4}$ University of Bonn}}%
\maketitle

\begin{abstract}
(Visual) Simultaneous Localization and Mapping (SLAM) remains a fundamental challenge in enabling autonomous systems to navigate and understand large-scale environments. Traditional SLAM approaches struggle to balance efficiency and accuracy, particularly in large-scale settings where extensive computational resources are required for scene reconstruction and Bundle Adjustment (BA). 
However, this scene reconstruction, in the form of sparse pointclouds of visual landmarks, is often only used within the SLAM system because navigation and planning methods require different map representations.
In this work, we therefore investigate a more scalable Visual SLAM (VSLAM) approach without reconstruction, mainly based on approaches for two-view loop closures. By restricting the map to a sparse keyframed pose graph without dense geometry representations, our `2GO' system achieves efficient optimization with competitive absolute trajectory accuracy. In particular, we find that recent advancements in image matching and monocular depth priors enable very accurate trajectory optimization without BA. 
We conduct extensive experiments on diverse datasets, including large-scale scenarios, and provide a detailed analysis of the trade-offs between runtime, accuracy, and map size. Our results demonstrate that this streamlined approach supports real-time performance, scales well in map size and trajectory duration, and effectively broadens the capabilities of VSLAM for long-duration deployments to large environments.
\end{abstract}
\vspace{-1mm}

\IEEEpeerreviewmaketitle
\section{Introduction}
\label{sec:intro}

\begin{figure}[t]
    \centering
    \includegraphics[width=\linewidth]{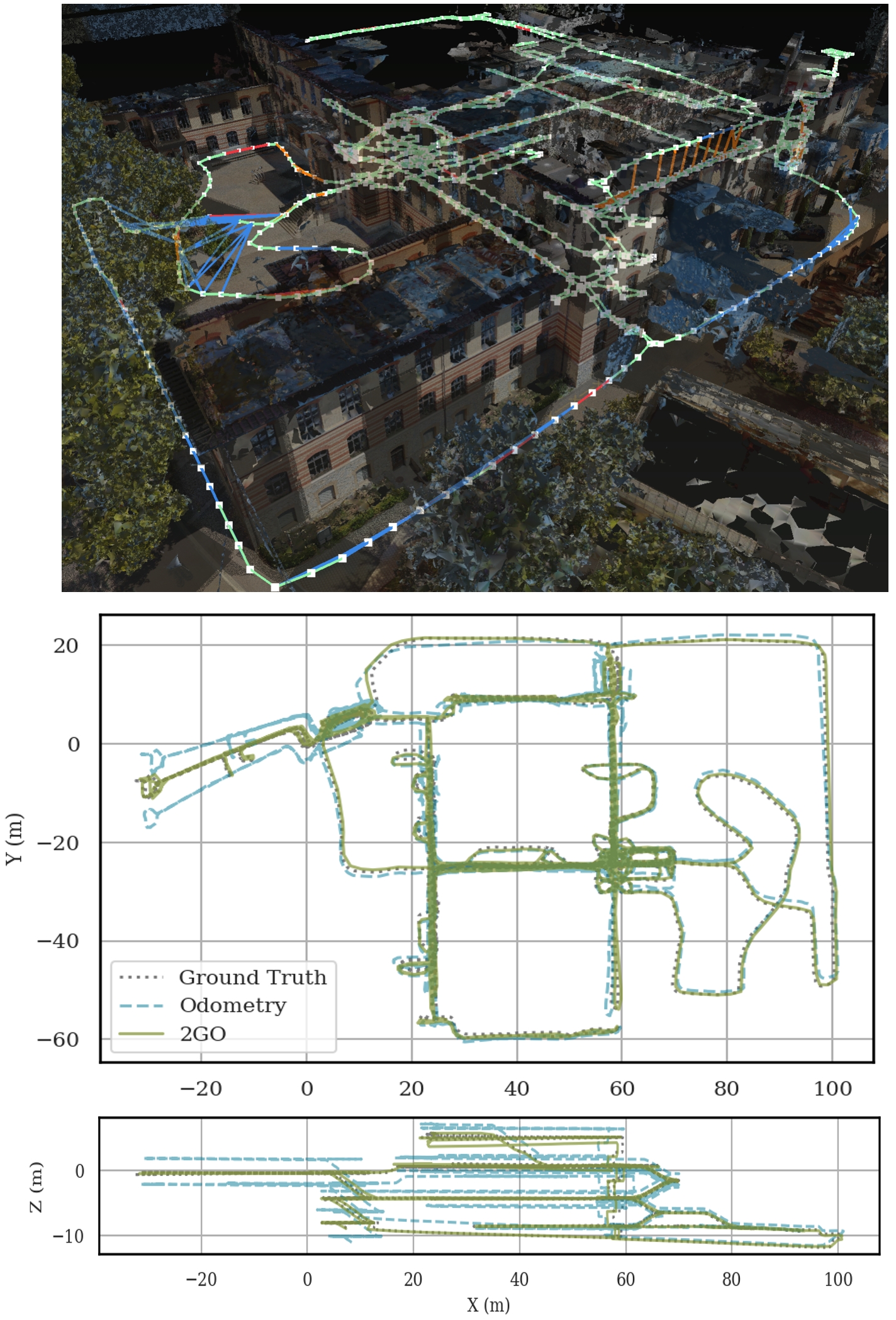}
    \caption{\textbf{Top}: 2GO, our proposed system, refines a \SI{1.1}{hr}, \SI{2.7}{km} trajectory collected from a quadrupedal robot, overlaid on a 3D reconstruction of the surroundings. Absolute and scale-free loop-closure edges are visible in yellow and blue, respectively. Loop-closure edges with both types of constraints present are in red. \textbf{Bottom}: 2GO significantly improves the quality of the input trajectory, particularly in the Z-direction.}
    \label{fig:teaser}
    \vspace{-1.\baselineskip}
\end{figure}

\label{sec:related_work}
\begin{figure*}[h]
    \centering
    \includegraphics[width=0.99\linewidth]{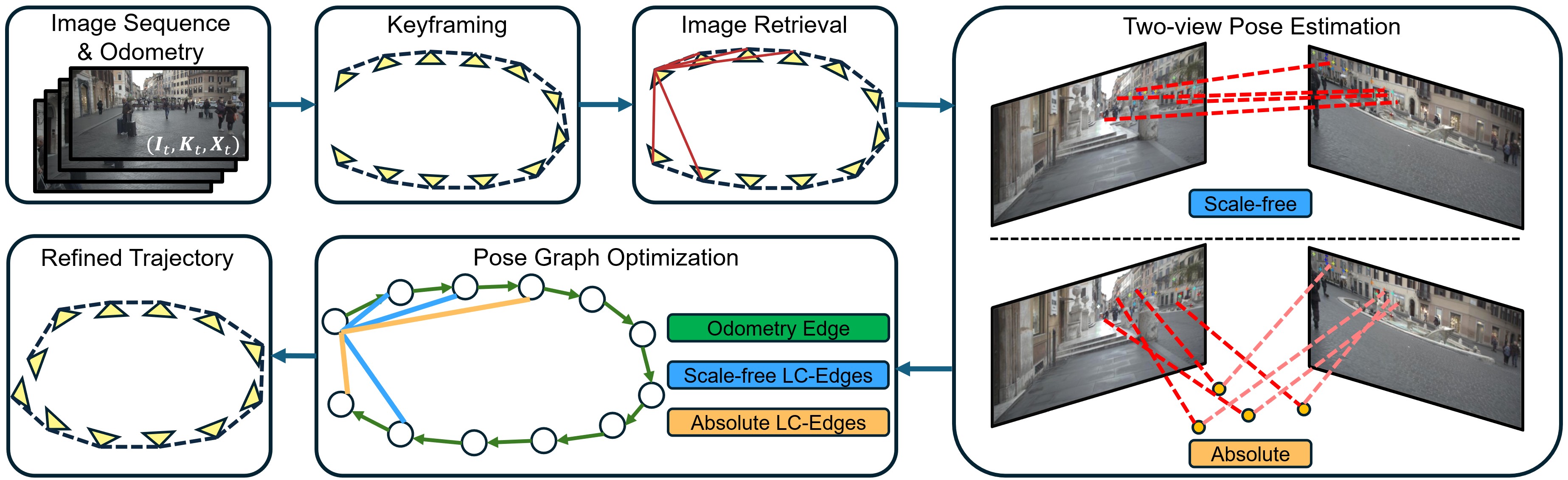}
    \caption{\textbf{System Overview.} From an input stream of odometry and images, we first downsample keyframes. Then we check for loop-closure candidates within previous keyframes and estimate the relative pose between candidate image pairs (metric and/or up-to-scale). Candidates are then filtered using geometric checks and then added as two-view edges to the pose-graph map, which is optimized with PGO. Marked in yellow is the main innovation of this paper.
    }
    \label{fig:system}
    \vspace{-1.\baselineskip}
\end{figure*}

Simultaneous Localization and Mapping (SLAM) methods have become a cornerstone in any autonomous robotic system. A robust, accurate, and scalable SLAM solution is crucial for applications that require navigation over large-scale environments and/or long deployment times. SLAM has been extensively studied over the past few decades~\cite{cadenda16past,ebadi2023present}, with methods deployed across various platforms in diverse environments for tasks such as robot navigation~\cite{fehr2018visual,werby2024hierarchical}, mixed reality~\cite{chen20243d,cruz2023mixed}, autonomous driving~\cite{andresen2020accurate,schaupp2020mozard}, and mobile device applications~\cite{baruch2021arkitscenes}. SLAM methods have become remarkably accurate even in very challenging environments~\cite{darpasubT}, but achieving scalability remains challenging due to the substantial computational and storage demands involved. 
VSLAM methods are of interest due to the widespread availability and low cost of camera sensors, but in competitions such as the HILTI SLAM Challenge~\cite{helmberger2022hilti}, their accuracy lags behind LiDAR SLAM. In summary, despite a long line of research, a VSLAM solution that is both scalable and accurate is still missing.

Concerning accuracy, there has been steady progress in VSLAM and VIO towards highly accurate pose estimates within a context of about \SI{10}{min}. Stereo visual odometry has even outperformed LiDAR-based methods on the KITTI benchmark~\cite{cvivsic2022soft2} (all trajectories in KITTI are shorter than \SI{10}{min}). Locally accurate odometry is 
available for visual-inertial setups~\cite{Qin_2018vinsmono, campos2021orb}, and monocular odometry makes steady progress~\cite{teed2023dpvo}. However, while visual odometry provides reliable local tracking, achieving globally accurate trajectory estimation requires loop closures and global optimization, where scalability remains challenging. State-of-the-art methods use pose graph optimization (PGO) or bundle adjustment (BA) for a globally consistent trajectory, with BA also providing a consistent reconstruction. PGO has a worst-case time complexity cubic in the number of poses~\cite{gtsam}, and the time complexity of BA is linear in the number of landmarks and cubic in the number of poses~\cite{agarwal2010bundle}. Consequently, as our experiments confirm, for large environments with many landmarks, and specifically for long trajectories with many poses, the optimization grows in complexity and takes too long to provide real-time updates.

This raises the question of how SLAM optimization complexity can be reduced. In particular, this work asks how far one can get without a 3D reconstruction.
The sparse landmark point cloud typically optimized by BA is rarely used in downstream robotic modules. It is often too sparse for operator visualization, and planning methods typically use other data structures, such as Truncated Signed Distance Fields (TSDF). In practice, fully-fledged mapping frameworks run separate processes to integrate a TSDF from the optimized poses~\cite{Cramariuc_2023maplab2}
. However, within existing SLAM frameworks, reconstructed landmarks are required to process loop closures and therefore maintaining map consistency over long horizons: Maplab~\cite{Cramariuc_2023maplab2} finds loop closures by matching landmarks that are close in descriptor space, applying geometric checks based on the reconstruction, and then merging matched landmarks to the same position. BA then optimizes the poses accordingly. ORB-SLAM3~\cite{campos2021orb} follows the same approach for loop closure optimization. 

In this work, we explore a more scalable approach to VSLAM based on two key insights: (i) robust and \textbf{locally accurate odometry} is widely available on most robotic systems and other devices; (ii) large deep-learning models are increasingly adept at estimating geometry from single and multiple views, providing useful vision priors that enable the construction of \textbf{global relations} between keyframes without requiring a map. While the first development reduces the margin of improvement for BA given a trajectory from odometry, the second point prompts us to reconsider if a global geometric map is necessary at all. Thus, we propose to optimize a globally consistent trajectory only through PGO, where loop-closure (LC) edges are constructed directly from pairs of keyframes. To this end, we revisit prior approaches for scale-free LC edges with modern image matchers and experiment with generating absolute LC constraints from metric monocular depth~\cite{hu2024metric3d} or the MASt3R two-view estimator~\cite{mast3r_arxiv24}.

Our 2-view PGO approach, which we call `2GO', therefore does not produce any 3D reconstruction. Instead, it uses locally accurate odometry for short-term consistency and vision priors to establish global relations between keyframes, enabling fully map-free optimization using PGO.
Through extensive experiments on large-scale trajectories, we show that this simplification yields significantly lower map size and runtime. Surprisingly, in most trajectories, 2GO also yields higher pose accuracy than state of the art VSLAM methods. In summary, our contributions are:
\begin{itemize}
    \item \textbf{2GO}, a two-view PGO framework that leverages the potential of recent advancements in monocular vision priors for computationally efficient and accurate VSLAM which scales to large-scale environments.
    \item Comprehensive analysis of runtime, accuracy, and map size of VSLAM methods on long trajectories and large scenes.
    \item Extensive experiments on large-scale datasets, with different odometry sources, and comparisons with a mapping system that performs global BA.
\end{itemize}

\section{Related Work}

\subsection{Large-Scale SLAM}

Conventional VSLAM and VIO approaches \cite{Qin_2018vinsmono,campos2021orb} build a metric 3D map and jointly optimize the map and camera pose. They have demonstrated reliable performance across various public benchmarks~\cite{Geiger2012CVPRkitti, Burri25012016euroc, helmberger2022hilti}. These methods achieve high local accuracy, but often face scalability challenges. To improve the scalability of VSLAM, previous methods have explored reducing the computational cost of BA by limiting it to a sliding window or running it at a low frequency~\cite{leutenegger2013keyframe, campos2021orb}. Some approaches eliminate BA entirely and rely solely on PGO, though they still require landmarks to establish loop closure constraints~\cite{Qin_2018vinsmono}.
In contrast, our approach not only skips BA but also eliminates the reconstruction and triangulation steps. 

\subsection{Vision-based Priors}
Recent SLAM methods have explored monocular depth to enhance accuracy and robustness. Early approaches, such as \cite{newcombe2011dtam}, optimize camera poses and dense depth maps using a regularized energy function. More recent Neural-SLAM methods, including \cite{Zhu2023NICER}, leverage monocular depth estimators for dense reconstruction. Predicting geometric priors from monocular inputs, such as depth and surface normals \cite{hu2024metric3d}, has made significant progress. However, these monocular priors remain underexplored in SLAM, particularly in sparse monocular settings. It remains unclear how to effectively use depth priors without performing dense reconstruction.

Beyond advances in monocular priors, two-view tasks have also seen significant progress, including feature matching \cite{sarlin20superglue, sun2021loftr}, visual place recognition (VPR) \cite{keetha2023anyloc}, and two-view reconstruction \cite{mast3r_arxiv24}. Improvements in feature detectors and matchers have enabled robust 2D-2D matching in challenging scenarios, such as wide-baseline settings. Global VPR methods have been widely explored and deployed in visual localization \cite{sarlin2019coarse}, with increasing attention on map-free localization due to their efficiency on edge devices, such as those used in augmented reality (AR) \cite{barroso2024mickey}. For instance, \textit{MARLoc} \cite{puigjaner2024augmentedrealitybordersachieving} integrates deep-learned 2D features and matchers with global image descriptors, achieving precise map-free localization on AR devices. A similar concept can be applied to SLAM to enhance loop-closure edge construction in a map-free manner. Two-view reconstruction priors have also been initially explored in both Structure-from-Motion (SfM) \cite{duisterhof2024mast3r} and VSLAM \cite{murai2024_mast3rslam}. However, these methods still rely on reconstruction maps and perform dense bundle adjustment.


\section{Loop Closures from Two Views}

\label{sec:methodology}
~\Cref{fig:system} provides an overview of 2GO. To constrain a trajectory using loop closures (LCs) without reconstructing 3D geometry, 2GO first finds LC candidates via image retrieval from previous keyframes and then employs different pose constraints between image pairs. Finally, pose graph optimization (PGO) is performed using the identified LC-Edges to refine the prior trajectory derived from odometry. In this section we introduce how we construct different variants of loop-closure edges from two views. In \Cref{sec:implementation} we describe how these two-view constraints are used in the proposed SLAM system.

\subsection{Problem Statement}

Given a sequence of images from a calibrated camera and their corresponding odometric poses, the input to 2GO at time $t$ is a 3-tuple $(\mathcal{I}_t, \mathbf{X}_t, \mathbf{K}_t)$. $\mathcal{I}_t \in \mathbb{R}^{H \times W \times C}$ is the rectified image from the camera. $\mathbf{X}_t \in \mathrm{SE(3)}$ is the pose of the camera with respect to a reference frame. And $\mathbf{K}_t \in \mathbb{R}^{3\times 3}$ are the camera intrinsics. 2GO incrementally refines the overall trajectory up to the present timestamp $N$, $\hat{\mathcal{X}} = ( \mathbf{\hat{X}}_1, \mathbf{\hat{X}}_2, \dots \mathbf{\hat{X}}_N )$. Without loss of generality, we describe the following methodology for a single-camera system. However, the approach extends naturally to multiple cameras, provided that the extrinsic transformations to the estimated odometry frame are known.

\begin{figure*}[h]
    \centering
    \includegraphics[width=0.99\linewidth]{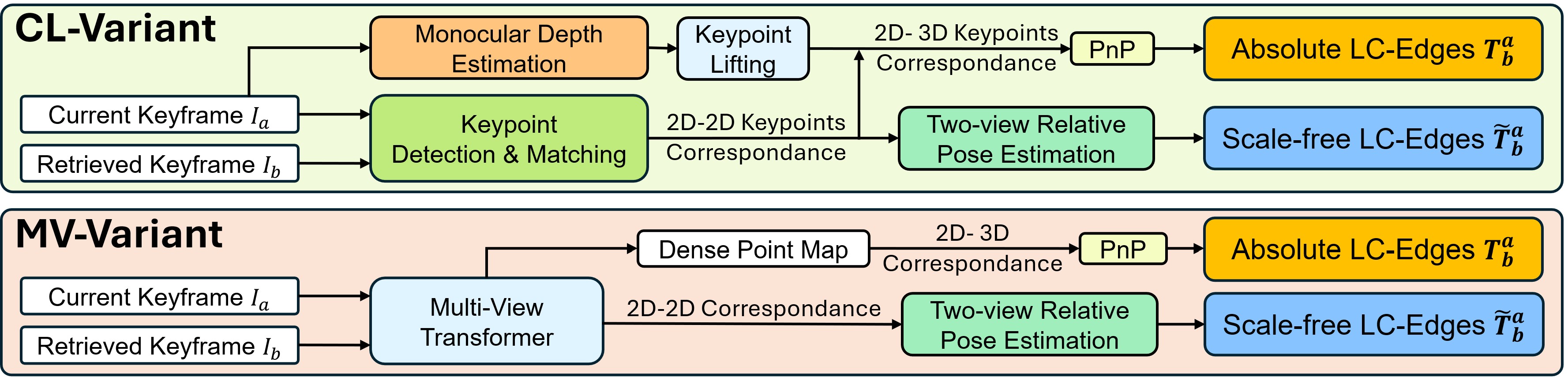}
    \caption{\textbf{Two-view Pose Estimation.} Both CL and MV variants establish 2D-2D and 2D-3D feature correspondences for scale-free and absolute LC-Edges, respectively. The CL-variant employs separate deep learning models in a modular pipeline, while the MV-variant achieves this with a single two-view reconstruction model.
    }
    \label{fig:two-view}
    \vspace{-1.\baselineskip}
\end{figure*}
\subsection{Image Retrieval}
\label{subsec:retrieval}
The image retrieval stage identifies LC candidates—image pairs that can establish two-view constraints by observing the same region of the environment. Given a new input keyframe $\mathcal{I}_k$, image retrieval searches previous keyframes $\mathcal{I}_{1,2,\dots,k-1}$ for LC candidates, according to their \textit{appearance} and \textit{geometric} similarity, analogous to the \textit{long-term} and \textit{mid-term} data associations described in ~\cite{campos2021orb}. Unlike conventional image retrieval methods for LC detection, 2GO prioritizes high recall over high precision, as multiple filtering stages (detailed in \Cref{sec:input-filtering-robustness}), mitigate outliers introduced during image retrieval. This aggressive strategy increases the likelihood of detecting valid LCs to refine the input trajectory.

\textbf{Similarity Candidates}: While many SLAM systems utilize bag-of-words based VPR~\cite{GalvezTRO12DBoW2}, only a few SLAM systems use global-descriptor based VPR approaches~\cite{cieslewski2018data}. These are more commonly used in SfM methods~\cite{sarlin2019coarse}, where image correspondences are retrieved from large collections of unordered images. 2GO leverages one such VPR model, \textit{BoQ}~\cite{Ali-bey_2024_CVPR}, which generates a global descriptor $\mathbf{F}_{t}$ for each image $\mathcal{I}_t$, to identify image pairs with visual similarity.

The VPR model is trained to ensure that the descriptors of two images are close in a similarity metric $\langle \mathbf{F}_a, \mathbf{F}_b \rangle$ (in our case the dot product), if $\mathcal{I}_a$ and $\mathcal{I}_b$ share visual overlap. Hence, we define a set of similarity LCs, $\mathcal{C}_{\text{sim}}$ for a given image $\mathcal{I}_k$ by selecting the $n_{\text{sim}}$ images with most similar descriptors:
\begin{align}
    \mathcal{C}_{k, \text{sim}} =  \left\lbrace \mathcal{I}_j\mid j \in \operatorname{Top}_{j=0...k-1}^{n_\text{sim}} \, \langle \mathbf{F}_k, \mathbf{F}_{j} \rangle  \right\rbrace 
\end{align}
where $\operatorname{Top}_A^n (\cdot)$ picks the top $n$ elements out of $A$ according to the given metric.









\textbf{Proximity Candidates}: We also consider candidate LCs based on the geometric difference between the current and previous keyframe poses. This allows retrieving images that might otherwise be missed due to low visual similarity, such as those affected by large viewpoint changes. Since the robust learned matchers~\cite{mast3r_arxiv24, lindenberger2023lightglue} used in the subsequent steps can establish correspondences even across wide baselines, incorporating such LC candidates is beneficial. To accomplish this, we first find the set of nearby previous keyframes:
\begin{align}
    \tilde{\mathcal{C}}_{k, \text{prox}} = \left\lbrace \mathcal{I}_j \mid \forall j : \text{Rot}(\hat{\mathbf{T}}^k_j) < \delta_r \land \text{Trans}(\hat{\mathbf{T}}^k_j) < \delta_t \right\rbrace
    \label{eqn:c_tilde_prox}
\end{align} 
where $\hat{\mathbf{T}}^k_j = {\hat{\mathbf{X}}_k}^{-1} {\hat{\mathbf{X}}_j}$ denotes the relative transform between the current estimate of the poses of keyframes $i$ and $j$, and $\delta_r$ and $\delta_t$ are the thresholds for rotational and translational differences, respectively. We prioritize matches with a larger time gap $\Delta t(\mathcal{I}_k, \mathcal{I}_j)$ between images, as they are more effective in correcting odometry drift.:
\begin{align}
    \mathcal{C}_{k, \text{prox}} =  \operatorname{Top}_{\mathcal{I}_j \in \tilde{\mathcal{C}}_{k, \text{prox}}}^{n_\text{prox}}  \Delta t(\mathcal{I}_k, \mathcal{I}_j) 
\end{align}

Subsequently, LC candidates $\mathcal{C} = \mathcal{C}_{\text{sim}} \cup \mathcal{C}_{\text{prox}}$ are passed to the two-view pose estimation module.

\subsection{Constructing Two-View LC-Edges}
\label{subsec:LC}
In conventional VSLAM and SfM methods, image poses are typically estimated via PnP using 2D-3D correspondences from an estimated metric map. Early Pose-SLAM approaches also relied on landmarks to define LC constraints without optimizing their positions~\cite{ila2009information}. 2GO eliminates the need for explicit 3D landmarks by (i) revisiting up-to-scale LC-edges from 2D-2D matches and (ii) leveraging metric monocular depth prior for single-view 3D geometry estimation.

For each keyframe $\mathcal{I}_k$, 2GO estimates its relative pose $\mathbf{T}^k_a \in \mathrm{SE(3)}$ to each LC candidate $\mathcal{I}_a \in \mathcal{C}$. \Cref{fig:two-view} illustrates two variants of this process for obtaining \textit{scale-free} $\tilde{\mathbf{T}}^k_a$ and \textit{absolute} $\mathbf{T}^k_a$ two-view LC-edges. The \textbf{CL} (\textbf{C}orrespond-\textbf{L}ift) variant follows conventional pipelines, using separate models for keypoint detection~\cite{tyszkiewicz2020disk}, matching~\cite{lindenberger2023lightglue}, and keypoint depth estimation via metric monocular depth (MDE)~\cite{hu2024metric3d}. In contrast, the \textbf{MV} (\textbf{M}ulti-\textbf{V}iew) variant employs a multi-view model such as MASt3R~\cite{mast3r_arxiv24}, which predicts 2D keypoint matches and 2D-3D correspondences through a single model.

\textbf{Scale-free LC-Edge:} The 2D keypoint correspondences between two images yield the essential matrix $\mathbf{E}$~\cite{Hartley:2003:MVG:861369}.
The up-to-scale relative pose between the two keyframes $\tilde{\mathbf{T}}^k_a = \begin{bmatrix} \mathbf{R}^k_a \mid\tilde{\mathbf{t}}^k_a \end{bmatrix}$ can then be obtained by decomposing $\mathbf{E} = \begin{bmatrix} \tilde{\mathbf{t}}^k_a \end{bmatrix}_\times \left(\mathbf{R}^k_a \right)^T$. Similar as~\cite{boler23essentialPoseSLAM}, the scale-free pose residual $\mathbf{r}_{S}(\hat{\mathbf{T}}^a_b, \tilde{\mathbf{T}}^a_b)$ is computed as the pose error between a scale-free pose measurement $\tilde{\mathbf{T}}^a_b$ and the expected measurement $\hat{\mathbf{T}}^a_b$ from the current estimate in the pose graph.
\begin{equation}
\mathbf{r}_{S}(\hat{\mathbf{T}}^a_b, \tilde{\mathbf{T}}^a_b) = \begin{bmatrix}
\tilde{\mathbf{t}}^a_b - \text{norm}({\hat{\mathbf{t}}^a_b}) \\
\mathbf{R}^a_b \boxminus \hat{\mathbf{R}}^a_b
\end{bmatrix} \in \mathbb{R}^6
\label{eq:scale-free-pose-factor}
\end{equation}
\begin{equation}
\text{norm}(\mathbf{t}) =
\begin{cases}
\frac{\mathbf{t}}{\|\mathbf{t}\|}, &  \|\mathbf{t}\| > 0, \\
\mathbf{0}, &  \|\mathbf{t}\| = 0.
\end{cases}
\end{equation}
Here, $\boxminus$ denotes the operation mapping the difference in expected and measured rotations from the Lie Group $\mathrm{SO(3)}$ to a vector in $\mathbb{R}^3$ for optimization~\cite{sola2018microlie}. In a scale-free pose measurement, the translation vector $\tilde{\mathbf{t}}^a_b$ is a unit vector. The expected translation $\hat{\mathbf{t}}^a_b$ is therefore also normalized before the two translation vectors are compared. The rare case $\|\tilde{\mathbf{t}}\| = 0$ is a pure rotation. To prevent numerical instabilities, these residuals are not included in the pose graph.



While one scale-free LC-Edge does not constrain the scale of the pose between two keyframes, they can still constrain the pose graph if enough scale-free edges are added. This is demonstrated in our experimental evaluations.


\textbf{Absolute LC-Edge:} In the CL-variant, a MDE model is used to obtain a metric depth map $\mathbf{D}_k \in \mathbb{R}^{H \times W}$ from $\mathcal{I}_k$. $\mathbf{D}_k$ is used to lift the detected keypoints $\mathbf{p}_0^k, \dots \mathbf{p}_N^k$ to 3D.
\begin{equation}
   \mathbf{P}_i^k = d_i^k K_k^{-1} \begin{bmatrix} \mathbf{p}_i^k \\1 \end{bmatrix}
\end{equation}
Here, $d_i^k$ denotes the estimated depth value at the pixel $\mathbf{p}_i^k$.

In the MV-variant, a model predicts a dense point-map $\mathbf{D'}\in\mathbb{R}^{H \times W \times 3}$, where each pixel in $\mathcal{I}_k$ has a predicted position relative to $\mathbf{X}_k$. The coordinates of $\mathbf{D'}$ indexed at $(u,v)=\mathbf{p}_i^k$, directly give $\mathbf{P}_i^k$.

Both two variants generate 3D-2D correspondences $(\mathbf{P}^k_i, \mathbf{p}^a_i)$ between $\mathcal{I}_k$ and $\mathcal{I}_a$. These allow a PnP solver to obtain an absolute pose between them, yielding $\mathbf{T}^k_a$, which is incorporated into the pose graph as an absolute LC-Edge. Some MV-Models can alternatively directly estimate $\mathbf{T}^k_a$~\cite{mapanything}.



    
    

\section{System Implementation}
\label{sec:implementation}
This section introduces how we build a complete trajectory estimation system around the two-view LC mechanism introduced in \Cref{sec:methodology}. Given the raw input image and odometry streams, we subsample the camera stream into keyframes. During the image retrieval as introduced in \Cref{subsec:retrieval}, we run filtering steps to rule out potential false-positive LCs. This loop is repeated for every new incoming keyframe and we construct the PGO that includes both filtered scale-free and absolute LC-edges to regularly optimize the trajectory.

\subsection{Keyframing}

2GO enforces spatial sparsity by choosing new keyframes solely based on the translational or rotational distance from the current frame to the previous keyframe, following a simplified subset of the rules from~\cite{Cramariuc_2023maplab2}. 2GO does not maintain a landmark map, making landmark triangulation between consecutive keyframes unnecessary. Hence, consecutive keyframes do not require a minimum number of covisible keypoints, allowing for a sparser selection strategy that enhances scalability. Furthermore, as relative poses between consecutive frames can be obtained from input odometry, sophisticated keyframing strategies based on covisible keypoints~\cite{Cramariuc_2023maplab2} or difference in optical flow~\cite{abate2024kimera2robustaccuratemetricsemantic} are not necessary.

\subsection{Two-View Pose Estimation}
To compute relative pose between two LC frames constructed in~\Cref{subsec:LC}, 2GO uses the \textit{COLMAP}~\cite{schoenberger2016sfm} implementation of relative pose estimation. For keypoint correspondences between images, the CL-variant pipeline uses \textit{DISK} features~\cite{tyszkiewicz2020disk} and the \textit{LightGlue}~\cite{lindenberger2023lightglue} matcher. We then estimate the metric depth through \textit{Metric3Dv2}~\cite{hu2024metric3d}. For the MV-Variant, we use \textit{Mast3r}~\cite{mast3r_arxiv24}. 2GO is not tightly coupled to using these choices. As newer models with better performance become available,  they can be swapped in easily. 

\subsection{LC Filtering} \label{sec:input-filtering-robustness}

Before integration into the pose graph, LC-Edges undergo feasibility filtering. This enhances robustness against false positives in image retrieval and minimizes redundant computation, in turn reducing overall runtime.

\textbf{Image Matching Feasibility}: Scale-free and absolute LC-Edges require sufficient 2D-2D and 2D-3D correspondences, respectively. This filter uses the image matchers to ensure that images have enough meaningful visual overlap.\\
\textbf{Geometric Feasibility}: During computation of scale-free and absolute LC-Edges, the respective solvers assess whether the given correspondences yield a consistent model. This ensures the overall geometric consistency of the correspondences from image matching.\\
\textbf{Pose-Graph Consistency}: The computed transform between images should remain consistent with the current optimized trajectory. This enhances robustness against false-positive matches in visually similar environments.


\subsection{Pose Graph Optimization}

Optimizing the pose graph is analogous to solving a nonlinear least squares problem. The \textit{iSAM2} incremental optimizer as implemented in \textit{GTSAM}~\cite{gtsam} is used as the back-end for optimization. In 2GO, the objective function of its PGO is formulated as:\vskip-20pt
\begin{equation}
\begin{split}
\hat{\mathcal{X}} = \argmin_{\mathcal{X}} \overbrace{ \| \mathbf{r}_{P}(\mathbf{X}_0) \|^2_{\Sigma_P} }^\text{prior} + \overbrace{
\sum_{e \in \mathcal{O}} \| \mathbf{r}_{A}(h(\mathcal{X}), e) \|^2_{\Sigma_e} }^\text{odometry edges} + \\
\underbrace{\sum_{l \in \mathcal{L}_a} \rho( \| \mathbf{r}_{A}(h(\mathcal{X}), l) \|^2_{\Sigma_{l_a}} )}_\text{absolute LCs} + \underbrace{ 
\sum_{l \in \mathcal{L}_s} \rho( \| \mathbf{r}_{S}(h(\mathcal{X}), l) \|^2_{\Sigma_{l_s}} )}_\text{scale-free LCs}
\end{split}
\label{eq:pgo-objective}
\end{equation}\vskip-5pt
where $\|\mathbf{a}\|^2_{\Sigma} = \mathbf{a}^T \Sigma \mathbf{a}$ denotes the L2 norm of $\mathbf{a} \in \mathbb{R}^n$ weighted by $\Sigma \in \mathbb{R}^{n \times n}$. $h(\mathcal{X})$ is the measurement function that obtains the relevant expected measurement from $\mathcal{X}$. $\rho(\cdot)$ denotes a robust Cauchy loss~\cite{lee2013robust}. The overall error sums over odometry edges, absolute LC-Edges, and scale-free LC-Edges. Three types of residual are defined:
\begin{itemize}
    \item $\mathbf{r}_{A}(\hat{\mathbf{T}}^a_b, \mathbf{T}^a_b) \in \mathbb{R}^6$, representing the pose error between an absolute pose measurement $\mathbf{T}^a_b \in \mathrm{SE(3)}$ from frame $b$ to $a$ and its expected measurement $\hat{\mathbf{T}}^a_b \in \mathrm{SE(3)}$ in the pose graph. This residual is well-studied and commonly used in PGO applications \cite{Qin_2018vinsmono, rosinol2020kimera}. The difference in expected and measured poses is mapped to a vector space $\mathbb{R}^6$ from $\mathrm{SE(3)}$ for optimization~\cite{sola2018microlie}.

    \item $\mathbf{r}_{P}(\mathbf{X}_0) = \mathbf{r}_{A}(\hat{\mathbf{X}}_0, \mathbf{X}_0) \in \mathbb{R}^6$ represents a prior, anchoring the current estimate of the first pose $\hat{\mathbf{X}}_0$ to its initial value $\mathbf{X}_0$.
    
    \item $\mathbf{r}_{S}(\hat{\mathbf{T}}^a_b, \tilde{\mathbf{T}}^a_b) \in \mathbb{R}^6$, is the pose error between a scale-free pose measurement $\tilde{\mathbf{T}}^a_b$ and the expected measurement $\hat{\mathbf{T}}^a_b$ from the current estimate in the pose graph, as ~\Cref{eq:scale-free-pose-factor} introduced in \Cref{subsec:LC}.
\end{itemize}

PGO is performed every $\lfloor N / l \rfloor$ keyframes, where $N$ is the total number of keyframes and $l$ is a tunable parameter. This keeps optimization tractable as the pose graph grows denser.

\section{Experiments}

\subsection{Experiment Setup}

2GO focuses on optimizing a trajectory from an odometry source in large-scale scenarios. Among the commonly used SLAM datasets, such as EuRoC~\cite{Burri25012016euroc}, HILTI-22~\cite{helmberger2022hilti}, and TUM-VI~\cite{schubert2018vidataset}, KITTI offers the longest trajectories with dense ground truth, making it a representative conventional benchmark. In addition, we select Vision Benchmark in Rome (VBR)~\cite{brizi2024vbrvisionbenchmarkrome}, a recent dataset with long-duration, large-scale sequences. To further demonstrate 2GO's real-world applicability, we evaluate it on a large-scale robot recording spanning \SI{1.1}{hr} and \SI{2.7}{km}. 

We use two VSLAM methods as baselines. \textbf{VINS-Fusion}~\cite{qin2019bvinsfusion} and \textbf{ORB-SLAM3}~\cite{campos2021orb} are both sliding-window methods with bag-of-words place recognition. When a loop closure is detected, VINS-Fusion performs 4-DOF PGO, while ORB-SLAM3 performs 
BA.
Both methods are run without IMU input. KITTI does not provide high-rate IMU data, and omitting IMU for VBR yielded better trajectories.

In addition, we compare 2GO to \textbf{Maplab2.0}~\cite{Cramariuc_2023maplab2}, a mapping framework capable of fusing multimodal sensor inputs. It can take an external odometry source as input and optimize its trajectory based on LCs and BA. It uses covisible visual keypoints for place recognition and performs loop closure by merging covisible keypoints between keyframes and performing global BA. We evaluate variants using BRISK~\cite{Leutenegger2011brisk} (\textbf{BIN}) and SuperPoint~\cite{detone2017superpoint} (\textbf{SP}) features. Unless otherwise stated, the CL-variant of 2GO is used, which is justified in \Cref{sec:result-lc-edges}.

\begin{table*}[h]
\centering
\resizebox{\linewidth}{!}{
\begin{tabular}{l|l|rr|r|rr|rr}
\toprule
Sequence & 2GO Variant & Median Trans. Err. & Median Rot. Err. & \#LC-Edges & ATE (m) $\downarrow$ & \% Decrease $\uparrow$ & Runtime (sec) $\downarrow$ & \% real-time $\downarrow$ \\
\midrule
\multirow{6}{*}{VBR \textit{ciampino\_train0}} 
& CL Scale-free  & \SI{0.86}{deg} & \SI{0.40}{deg} & 2722 & 28.54 & 78.27          &   932.07 &  60.17 \\
& CL Absolute    &   \SI{0.25}{m} & \SI{0.49}{deg} & 3403 & 25.12 & 80.87          & 2,067.31 & 133.47 \\
& CL Both        & /              & /              & 6308 & 24.69 & 81.20          & 2,086.56 & 134.71 \\
& MV Scale-free  & \SI{0.83}{deg} & \SI{0.38}{deg} & 1598 & 23.99 & \textbf{81.73} & 1,827.37 & 117.98 \\
& MV Absolute    &   \SI{2.12}{m} & \SI{2.56}{deg} & 3349 & 24.65 & 81.23          & 1,773.44 & 114.49 \\
& MV Both        & /              & /              & 4914 & 24.76 & 81.14          & 1,785.71 & 115.29 \\
\midrule
\multirow{6}{*}{KITTI-00} 
& CL Scale-free  & \SI{1.59}{deg} & \SI{0.33}{deg} &  426 & 2.77  &  78.87          & 198.99 & 42.39 \\
& CL Absolute    & \SI{0.86}{m}   & \SI{0.64}{deg} &  517 & 1.47  &  88.74          & 387.16 & 82.48 \\
& CL Both        & /              & /              & 1286 & 1.40  &  \textbf{89.33} & 389.04 & 82.88 \\
& MV Scale-free  & \SI{1.30}{deg} & \SI{0.55}{deg} &   61 & 14.83 & -13.30          & 201.76 & 42.98 \\
& MV Absolute    & \SI{4.82}{m}   & \SI{1.02}{deg} &  360 & 2.25  &  82.83          & 229.36 & 48.86 \\
& MV Both        & /              & /              &  436 & 2.20  &  83.18          & 227.06 & 48.37 \\
\bottomrule
\end{tabular}
}
\caption{\textbf{Performance comparison of different variants and LC-Edge combinations of 2GO.} VINS-Fusion odometry is used for both sequences. The best performing 2GO variant for each dataset is \textbf{bolded}.
For \textit{ciampino\_train0}, RMSE ATE: \SI{131.33}{m}, real-time duration: \SI{1549}{sec}.
For KITTI-00, RMSE ATE: \SI{13.09}{m}, real-time duration: \SI{469}{sec}. }
\label{tab:LC-Edge-ablation-merged}
\vspace{-1.5\baselineskip}
\end{table*}

\subsection{Is 2GO more scalable?} 

We first show 2GO's runtime and storage scalability in large-scale scenarios. \Cref{tab:mapsize-runtime-comparison} shows the runtime and storage size of different methods on the longest handheld VBR sequence, \textit{spagna\_train0}, with a distance of \SI{1.56}{km} and a \SI{23}{min} \SI{34}{sec} duration. 2GO only requires $0.83\,\times$ of the real-time duration to process the entire input sequence. The only other method within real-time is Maplab-BIN, while ORB-SLAM3 and VINS-Fusion run close to real-time.

Two baselines are much slower: VINS-Fusion performing PGO takes $4.4\,\times$ real-time to iteratively perform global PGO, as it uses the same keyframing strategy for odometry and pose refinement, resulting in more than 10000 keyframes. In contrast, 2GO optimizes 740 keyframes. Though both methods do not perform BA, this illustrates the growing time complexity of PGO with the number of input poses. The second outlier, Maplab-SP, requires about $10\,\times$ real-time to generate a map of camera poses and landmarks, as it uses the computationally expensive SuperPoint and SuperGlue models to detect and match keypoints between all consecutive frames. 

Conversely, Maplab-BIN uses lightweight BRISK features, completing its mapping stage in $0.08\times$ real-time. However, these features are less discriminative and less robust. As fewer loop closures are found, the optimization duration is shorter. Therefore, it cannot reduce the input odometry's ATE.

We compare map sizes between Maplab, ORB-SLAM3 and 2GO, as VINS-Fusion lacks map-saving functionality. The maps saved by Maplab and ORB-SLAM3 minimally include the pose graph, as well as the landmark locations, descriptors, and keypoint coordinates in each keyframe. In contrast, 2GO saves a pose graph, as well as a compressed image and a global descriptor for each keyframe. 2GO needs less than half of the storage of Maplab-BIN, a quarter of Maplab-SP, and a tenth of ORB-SLAM3. However, none of the maps are prohibitively large considering the trajectory duration.

This shows that 2GO scales well to long trajectories, even while yielding the lowest trajectory error. It remains one of the few methods within real-time constraints while maintaining the smallest map representation among all baselines.

\begin{table}[t]
\centering
\resizebox{\linewidth}{!}{
\begin{tabular}{c|c|c|c|c|c|c}
\toprule
Variant & \multicolumn{1}{c|}{VINS-} & \multicolumn{1}{c|}{VINS-} & ORB- & \multicolumn{1}{c|}{Maplab-BIN} & \multicolumn{1}{c|}{Maplab-SP} & 2GO \\
        & \multicolumn{1}{c|}{Fusion} & \multicolumn{1}{c|}{PGO} & SLAM3 &  & & (Ours) \\
\midrule
Runtime (sec) $\downarrow$ & 1564  
              & 6206 
              & 1755
              & \multicolumn{1}{c|}{\begin{tabular}[c]{@{}c@{}}\textbf{110} (Map)\\ \textbf{29} (Opt)\end{tabular}} 
              & \multicolumn{1}{c|}{\begin{tabular}[c]{@{}c@{}}14139 (Map)\\ 271 (Opt)\end{tabular}} 
              & 1167 \\
\midrule
Map Size (MB) $\downarrow$ & /  
              & /  
              & 1136
              & \multicolumn{1}{c|}{319}  
              & \multicolumn{1}{c|}{572}  
              & \textbf{146} \\
\midrule
RMSE ATE (m) $\downarrow$  & 21.14 
              & 13.22
              & 5.62
              & 20.83
              & 20.35
              & \textbf{3.73}               \\
\bottomrule
\end{tabular}}
\caption{\textbf{Comparison of runtime and map size.} Result across different methods for the \textit{spagna\_train0} trajectory of VBR (1414s real-time duration). ``VINS-PGO'' denotes VINS-Fusion with loop closure enabled. ``Maplab-BIN'' and ``Maplab-SP'' denote Maplab pipeline using Binary feature with nearest neighbor matcher and SuperPoint feature with SuperGlue matcher.}
\label{tab:mapsize-runtime-comparison}
\vspace{-1.5\baselineskip}
\end{table}

\subsection{How do different LC-Edges contribute to the optimization?}
\label{sec:result-lc-edges}
We assess 2GO's performance across CL/MV variants and different LC-Edge combinations. \Cref{tab:LC-Edge-ablation-merged} shows results on the KITTI-00 and \textit{ciampino\_train0} sequences with VINS-Fusion odometry.
LC-Edge accuracy is measured by comparing the estimated transformation $\mathbf{T}^k_a$ between two keyframes $k$ and $a$ with the transform between their ground-truth poses, 
${\mathbf{X}^\star_k}^{-1} {\mathbf{X}^\star_a}$. The translation error for Scale-Free LC-Edges is the angle between the normalized unit vectors $\tilde{\textbf{t}}^k_a$ and $\text{norm}({\textbf{t}^k_a}^{\star})$.

CL and MV variants using Absolute and Scale-Free (``Both'') LC-Edges perform well on both sequences. Scale-Free LC-Edge accuracy is similar across both variants, while Absolute LC-Edges from CL are much more accurate. The CL variant finds more LC-Edges. In addition, Scale-Free LC-Edges can refine trajectories, with both variants using only these LC-Edges reducing the ATE except for MV with KITTI-00, due to the low number of LC-Edges found.

In terms of runtime, where Absolute LC-Edges are used, MV variants run faster, perhaps because MASt3R performs both keypoint matching and depth estimation jointly. However, when only Scale-Free LC-Edges are used, the CL does not require MDE, and thus enjoys much lower runtime. 

Overall, because the CL Both variant finds many accurate LC-Edges, it is likely to be more robust across varying scenes, and thus is used for all subsequent experiments.

\begin{table*}[t]
\label{sec:experimental_results}
\footnotesize
\centering

\newcommand{\noimu}{{\scriptsize no imu}}
\newcommand{\novio}{{\scriptsize vio failed}}

\resizebox{\linewidth}{!}{
\setlength{\tabcolsep}{2pt}
\begin{tabular}{lrr|rrrrr|rrrrr}
\toprule
\multirow{2}{*}{Sequence} & \multirow{2}{*}{Length} & \multirow{2}{*}{Duration} & \multicolumn{5}{c|}{w/ ORB-SLAM3 Odometry} & \multicolumn{5}{c}{w/ VINS-Fusion Odometry} \\
&&                                              & odom only & ORB-SLAM3 & Maplab-BIN & Maplab-SP & 2GO   & odom only & VINS-PGO & Maplab-BIN & Maplab-SP & 2GO   \\
\midrule
KITTI-00 & \SI{3.72}{km} & \SI{470}{sec}            & 4.20       & 4.85      &   37.51    &  12.69         & \textbf{1.82}  & 13.09 & 4.20 & 11.13      & 11.39     & \textbf{1.58}  \\
KITTI-01 & \SI{2.45}{km} & \SI{114}{sec}           & 12.71       & 13.62     &   116.91         &  57.40         & \textbf{11.00} & 7.95 & 7.98 & 66.31      & 20.92     & \textbf{7.93}  \\
KITTI-02 & \SI{5.07}{km} & \SI{483}{sec}          & 8.05      & \textbf{5.64}      &   18.01    &   44.78   & 8.04  & 21.02 & 13.03 & 23.96      & 18.29     & \textbf{10.58} \\
KITTI-03 & \SI{0.56}{km} & \SI{83}{sec}          & 1.29       & 1.38      &    4.31    &   5.68    & \textbf{1.12}  & 1.64 & 1.55 &  4.04       & 4.31      & \textbf{1.17}  \\
KITTI-04 & \SI{0.39}{km} & \SI{28}{sec}          & \textbf{0.24}       & 0.34      &    0.93    &   2.65    & 0.28  & 1.30 & \textbf{1.25} & 2.27       & 2.47      & 1.31  \\
KITTI-05 & \SI{2.21}{km} & \SI{288}{sec}         & 2.05       & 0.95      &    7.44   &   7.98    & \textbf{0.92}  & 6.39 & 3.68 & 11.37      & 11.21     & \textbf{1.70}  \\
KITTI-06 & \SI{1.23}{km} & \SI{114}{sec}         & 2.05       & 0.79      &    7.09   &   10.74   & \textbf{0.62}  & 3.54 & 1.68 & 13.27      & 14.76     & \textbf{1.48}  \\
KITTI-07 & \SI{0.69}{km} & \SI{114}{sec}         & 1.19       & 0.50      &    5.60   &   5.63    & \textbf{0.41}  & 2.11 & 0.65 & 7.48       & 7.70      & \textbf{0.46}  \\
KITTI-08 & \SI{3.22}{km} & \SI{423}{sec}         & \textbf{3.50}       & 3.67      &    40.79  &   11.65   & 3.78  & 10.43 & 10.01 & 17.04      & 14.27     & \textbf{4.03}  \\
KITTI-09 & \SI{1.71}{km} & \SI{165}{sec}         & 2.87       & 3.22      &    10.23  &   11.32   & \textbf{1.84}  & 7.80 & 7.71 & 15.97      & 12.31     & \textbf{4.66}  \\
KITTI-10 & \SI{0.92}{km} & \SI{125}{sec}         & 1.31       & 1.19      &    44.09  &   11.05   & \textbf{0.91}  & 3.77 & 3.73 & 9.17       & 9.46      & \textbf{1.57}  \\
\midrule
campus\_train0 & \SI{2.73}{km} & \SI{602}{sec}   & 28.92      & 11.94         &    30.73  &   28.22   & \textbf{9.32} & 47.74  & 43.84  &  49.46  & 46.28   & \textbf{12.30}  \\
campus\_train1 & \SI{2.95}{km} & \SI{584}{sec}   & 28.56      & \textbf{7.99} &    30.88  &   27.42   & 8.98          & 46.94 & 23.40  &  44.43  & 34.93   & \textbf{9.60}  \\
ciampino\_train0 & \SI{9.01}{km} & \SI{1549}{sec} & 82.82     & 68.11         &    83.89  &   81.77   & \textbf{18.29} & 131.33 & 95.95  &  130.20 & 130.94  & \textbf{22.84} \\
ciampino\_train1 & \SI{5.20}{km} & \SI{942}{sec}  & 25.98     & 13.83         &    27.61  &   25.00   & \textbf{13.25} & 47.21  & 34.68  &  40.94  & 40.79   & \textbf{14.20}  \\
colosseo\_train0 & \SI{1.45}{km} & \SI{882}{sec}  & 11.09     & 9.67          &    9.40   &   10.29   & \textbf{6.80}  & 21.49  & 13.37  & 21.48   & 21.47   & \textbf{6.00}  \\
diag\_train0 & \SI{1.04}{km} & \SI{1003}{sec}     & /        & /              & /         & /         & /              & 27.35 & 33.10 & 27.17  & 27.23 & \textbf{26.70}  \\
pincio\_train0 & \SI{1.28}{km} & \SI{1114}{sec} & 6.30        & \textbf{2.08} &    6.31   &   6.91    & 2.79        & 11.29 & 7.75  & 11.22 & 11.23 & \textbf{3.18}  \\
spagna\_train0 & \SI{1.56}{km} & \SI{1414}{sec} & 11.42       & \textbf{5.62} &    11.95  &   12.01   & 6.48  & 21.14 & 13.22 & 20.83 & 20.35 & \textbf{3.73}  \\
\bottomrule
\end{tabular}}

\caption{RMSE ATE (m) after 6-DOF trajectory alignment.
``VINS-PGO'' denotes VINS-Fusion with loop closure enabled. ``Maplab-BIN'' denotes Maplab using Binary features with nearest neighbor matcher. ``Maplab-SP'' denotes Maplab using SuperPoint features with SuperGlue matcher.
The best method for each odometry source is \textbf{bolded}.
As mentioned in \Cref{sec:result-lc-edges}, the CL-Both variant of 2GO is used.
ORB-SLAM3 lost visual tracking on the \textit{diag\_train0} VBR sequence.
}
\label{tab:overall-result}
\vspace{-1.\baselineskip}
\end{table*}

\subsection{How accurate is 2-View PGO?}
We evaluate global trajectory error by comparing the RMSE Absolute Trajectory Error (ATE),
computed after applying a 6-DOF alignment
between the input and ground-truth trajectories. Results are shown in \Cref{tab:overall-result}. 
To ensure fair evaluation of Maplab and 2GO as trajectory refinement methods, the loop-closing functionality of ORB-SLAM3 is disabled when used as an odometry source. In sumamry, 2GO consistently improves upon input odometry in nearly all cases and often achieves the highest accuracy among the compared PGO and BA-based methods. 

On the long trajectories in VBR, 2GO strongly outperforms VINS-PGO and Maplab in refining the same input odometry, showing the effectiveness of integrating learned priors in PGO. While the performance gap between 2GO and ORB-SLAM3 performing BA is smaller, long, loopy sequences like \textit{ciampino\_train0} emphasize 2GO's ability to capitalize on loop closures for trajectory refinement. Several sequences in KITTI do not contain loops, so 2GO cannot improve upon input odometry as significantly.

\begin{figure}[t]
    \centering
    \includegraphics[width=0.95\linewidth]{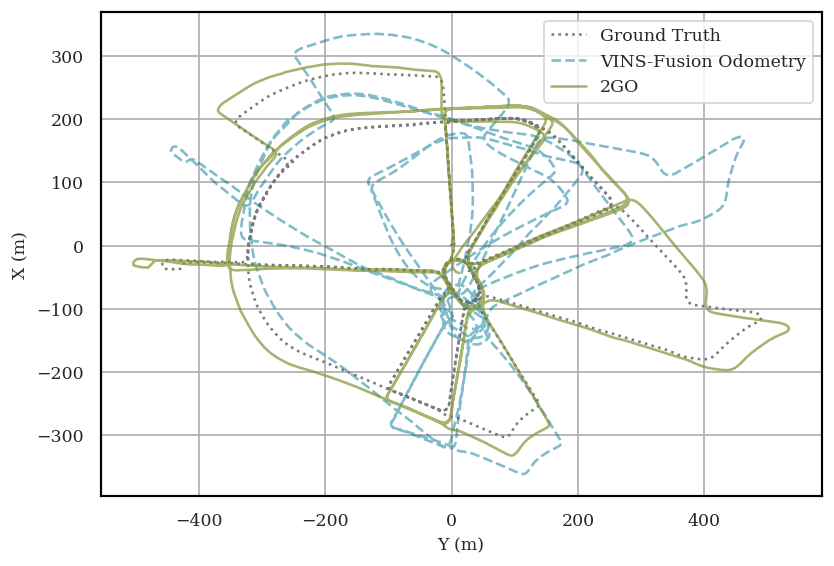}
    \caption{\textit{ciampino\_train0} sequence: 2GO improves on VINS-Fusion odometry, reducing ATE from \SI{131.33}{m} to \SI{22.84}{m}.}
    \label{fig:ciampino_train0-vinsFusion-refine}
\end{figure}

\begin{figure}[t]
    \centering
    \includegraphics[width=0.99\linewidth]{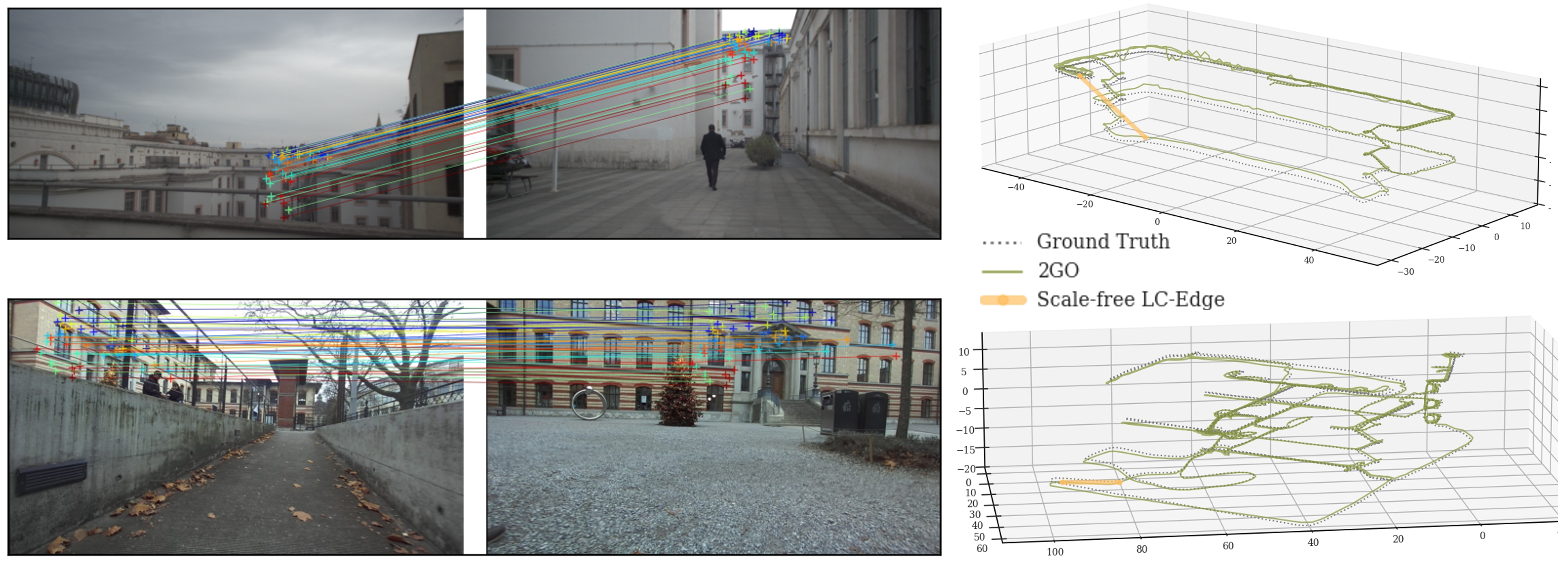}
    \caption{\textbf{Scale-free LC Edge Examples} The top row contains samples from \textit{diag\_train0}, while the bottom row is from our real-world recording. Both LCs are constructed from image pairs with large baselines and contribute to trajectory refinement.}
    \label{fig:lc_example}
\end{figure}

\Cref{fig:ciampino_train0-vinsFusion-refine} illustrates 2GO refining VINS-Fusion odometry on \textit{ciampino\_train0}, reducing ATE from \SI{131.33}{m} to \SI{22.84}{m}. \Cref{fig:lc_example} shows an example of 2GO constructing and leveraging loop-closure edges from image pairs with significantly different viewpoints. 

\begin{table}[]
\resizebox{\linewidth}{!}{
\centering
\begin{tabular}{l|rrrrrr}
\toprule
        & Spot Odom & ZED Odom & ZED w/ LC & 2GO   \\
\midrule
RMSE ATE (m)  & 2.44 & 26.18 & 23.05 & \textbf{0.66}  \\
\bottomrule
\end{tabular}}
\caption{\textbf{Comparison of mapping methods on a real-world trajectory.} Spot Odom is used as odometry input for 2GO. Over a \SI{1.1}{hr} real-world trajectory, 2GO requires only \SI{15.0}{min} for optimization, and saves a \SI{162.5}{MB} map.}
\label{tab:cab-traj-cmp}
\vspace{-2.0\baselineskip}
\end{table}

\subsection{How does 2GO work on a real-world robot?}
We assess the accuracy, scalability and real-world applicability of 2GO with a \SI{1.1}{hr}, \SI{2.7}{km} data recording on a Boston Dynamics Spot robot. The trajectory primarily covers different areas of a multi-floor building and a short outdoor segment, as shown in~\Cref{fig:teaser}. A Stereolabs Zed 2 camera provides images, IMU, and on-device VIO. Spot also provides device odometry, fused from leg and visual sources. 

We use the pipeline of~\cite{sarlin2022lamar} to generate ground truth poses. 
Our results in \Cref{tab:cab-traj-cmp} confirm that 2GO can be readily deployed on existing platforms using their provided odometry. It consistently operates within real-time constraints while maintaining a low-memory footprint for the map.

\section{Conclusion}
\label{sec:conclusion}

In this work, we presented 2GO, a novel approach to scalable visual SLAM that eliminates the need for dense scene reconstruction by leveraging two-view loop closures and monocular priors. By constraining the map to a sparse keyframed pose graph, we achieve a streamlined and computationally efficient optimization process while maintaining competitive trajectory accuracy. Our experiments on diverse datasets, including large-scale scenarios, highlight the effectiveness of recent advancements in image matching and monocular metric depth estimation for accurate trajectory estimation. The results demonstrate that 2GO enables real-time performance, scales effectively in both map size and trajectory duration, and is well-suited for long-duration deployments in large environments. This work broadens the scope of visual SLAM by offering a practical alternative for scenarios where dense reconstruction is unnecessary, paving the way for efficient and scalable solutions in autonomous navigation and mapping.




\bibliographystyle{IEEEtran}
\bibliography{references}

\begin{thebibliography}{10}
\providecommand{\url}[1]{#1}
\csname url@samestyle\endcsname
\providecommand{\newblock}{\relax}
\providecommand{\bibinfo}[2]{#2}
\providecommand{\BIBentrySTDinterwordspacing}{\spaceskip=0pt\relax}
\providecommand{\BIBentryALTinterwordstretchfactor}{4}
\providecommand{\BIBentryALTinterwordspacing}{\spaceskip=\fontdimen2\font plus
\BIBentryALTinterwordstretchfactor\fontdimen3\font minus \fontdimen4\font\relax}
\providecommand{\BIBforeignlanguage}[2]{{%
\expandafter\ifx\csname l@#1\endcsname\relax
\typeout{** WARNING: IEEEtran.bst: No hyphenation pattern has been}%
\typeout{** loaded for the language `#1'. Using the pattern for}%
\typeout{** the default language instead.}%
\else
\language=\csname l@#1\endcsname
\fi
#2}}
\providecommand{\BIBdecl}{\relax}
\BIBdecl

\bibitem{cadenda16past}
\BIBentryALTinterwordspacing
C.~Cadena, L.~Carlone, H.~Carrillo, Y.~Latif, D.~Scaramuzza, J.~Neira, I.~D. Reid, and J.~J. Leonard, ``Simultaneous localization and mapping: Present, future, and the robust-perception age,'' \emph{CoRR}, vol. abs/1606.05830, 2016. [Online]. Available: \url{http://arxiv.org/abs/1606.05830}
\BIBentrySTDinterwordspacing

\bibitem{ebadi2023present}
K.~Ebadi, L.~Bernreiter, H.~Biggie, G.~Catt, Y.~Chang, A.~Chatterjee, C.~E. Denniston, S.-P. Desch{\^e}nes, K.~Harlow, S.~Khattak \emph{et~al.}, ``Present and future of slam in extreme environments: The darpa subt challenge,'' \emph{IEEE Transactions on Robotics}, 2023.

\bibitem{fehr2018visual}
M.~Fehr, T.~Schneider, and R.~Siegwart, ``Visual-inertial teach and repeat powered by google tango,'' in \emph{2018 IEEE/RSJ International Conference on Intelligent Robots and Systems (IROS)}.\hskip 1em plus 0.5em minus 0.4em\relax IEEE, 2018, pp. 1--9.

\bibitem{werby2024hierarchical}
A.~Werby, C.~Huang, M.~B{\"u}chner, A.~Valada, and W.~Burgard, ``Hierarchical open-vocabulary 3d scene graphs for language-grounded robot navigation,'' in \emph{First Workshop on Vision-Language Models for Navigation and Manipulation at ICRA 2024}, 2024.

\bibitem{chen20243d}
J.~Chen, B.~Sun, M.~Pollefeys, and H.~Blum, ``A 3d mixed reality interface for human-robot teaming,'' in \emph{2024 IEEE International Conference on Robotics and Automation (ICRA)}.\hskip 1em plus 0.5em minus 0.4em\relax IEEE, 2024, pp. 11\,327--11\,333.

\bibitem{cruz2023mixed}
C.~Cruz~Ulloa, J.~del Cerro, and A.~Barrientos, ``Mixed-reality for quadruped-robotic guidance in sar tasks,'' \emph{Journal of Computational Design and Engineering}, vol.~10, no.~4, pp. 1479--1489, 2023.

\bibitem{andresen2020accurate}
L.~Andresen, A.~Brandemuehl, A.~Honger, B.~Kuan, N.~V{\"o}disch, H.~Blum, V.~Reijgwart, L.~Bernreiter, L.~Schaupp, J.~J. Chung \emph{et~al.}, ``Accurate mapping and planning for autonomous racing,'' in \emph{2020 IEEE/RSJ international conference on intelligent robots and systems (IROS)}.\hskip 1em plus 0.5em minus 0.4em\relax IEEE, 2020, pp. 4743--4749.

\bibitem{schaupp2020mozard}
L.~Schaupp, P.~Pfreundschuh, M.~B{\"u}rki, C.~Cadena, R.~Siegwart, and J.~Nieto, ``Mozard: Multi-modal localization for autonomous vehicles in urban outdoor environments,'' in \emph{2020 IEEE/RSJ International Conference on Intelligent Robots and Systems (IROS)}.\hskip 1em plus 0.5em minus 0.4em\relax IEEE, 2020, pp. 4828--4833.

\bibitem{baruch2021arkitscenes}
G.~Baruch, Z.~Chen, A.~Dehghan, T.~Dimry, Y.~Feigin, P.~Fu, T.~Gebauer, B.~Joffe, D.~Kurz, A.~Schwartz \emph{et~al.}, ``Arkitscenes: A diverse real-world dataset for 3d indoor scene understanding using mobile rgb-d data,'' \emph{arXiv preprint arXiv:2111.08897}, 2021.

\bibitem{darpasubT}
\BIBentryALTinterwordspacing
T.~Rou\v{c}ek, M.~Pecka, P.~\v{C}\'{\i}\v{z}ek, T.~Pet\v{r}\'{\i}\v{c}ek, J.~Bayer, V.~\v{S}alansk\'{y}, D.~He\v{r}t, M.~Petrl\'{\i}k, T.~B\'{a}\v{c}a, V.~Spurn\'{y}, F.~Pomerleau, V.~Kubelka, J.~Faigl, K.~Zimmermann, M.~Saska, T.~Svoboda, and T.~Krajn\'{\i}k, ``Darpa subterranean challenge: Multi-robotic exploration of underground environments,'' in \emph{Modelling and Simulation for Autonomous Systems: 6th International Conference, MESAS 2019, Palermo, Italy, October 29–31, 2019, Revised Selected Papers}.\hskip 1em plus 0.5em minus 0.4em\relax Berlin, Heidelberg: Springer-Verlag, 2019, p. 274–290. [Online]. Available: \url{https://doi.org/10.1007/978-3-030-43890-6_22}
\BIBentrySTDinterwordspacing

\bibitem{helmberger2022hilti}
M.~Helmberger, K.~Morin, B.~Berner, N.~Kumar, G.~Cioffi, and D.~Scaramuzza, ``The hilti slam challenge dataset,'' \emph{IEEE Robotics and Automation Letters}, vol.~7, no.~3, pp. 7518--7525, 2022.

\bibitem{cvivsic2022soft2}
I.~Cvi{\v{s}}i{\'c}, I.~Markovi{\'c}, and I.~Petrovi{\'c}, ``Soft2: Stereo visual odometry for road vehicles based on a point-to-epipolar-line metric,'' \emph{IEEE Transactions on Robotics}, vol.~39, no.~1, pp. 273--288, 2022.

\bibitem{Qin_2018vinsmono}
\BIBentryALTinterwordspacing
T.~Qin, P.~Li, and S.~Shen, ``Vins-mono: A robust and versatile monocular visual-inertial state estimator,'' \emph{IEEE Transactions on Robotics}, vol.~34, no.~4, p. 1004–1020, Aug. 2018. [Online]. Available: \url{http://dx.doi.org/10.1109/TRO.2018.2853729}
\BIBentrySTDinterwordspacing

\bibitem{campos2021orb}
C.~Campos, R.~Elvira, J.~J.~G. Rodr{\'\i}guez, J.~M. Montiel, and J.~D. Tard{\'o}s, ``Orb-slam3: An accurate open-source library for visual, visual--inertial, and multimap slam,'' \emph{IEEE Transactions on Robotics}, vol.~37, no.~6, pp. 1874--1890, 2021.

\bibitem{teed2023dpvo}
Z.~Teed, L.~Lipson, and J.~Deng, ``Deep patch visual odometry,'' \emph{Advances in Neural Information Processing Systems}, 2023.

\bibitem{gtsam}
\BIBentryALTinterwordspacing
F.~Dellaert and G.~Contributors, ``borglab/gtsam,'' May 2022. [Online]. Available: \url{https://github.com/borglab/gtsam)}
\BIBentrySTDinterwordspacing

\bibitem{agarwal2010bundle}
S.~Agarwal, N.~Snavely, S.~M. Seitz, and R.~Szeliski, ``Bundle adjustment in the large,'' in \emph{Computer Vision--ECCV 2010: 11th European Conference on Computer Vision, Heraklion, Crete, Greece, September 5-11, 2010, Proceedings, Part II 11}.\hskip 1em plus 0.5em minus 0.4em\relax Springer, 2010, pp. 29--42.

\bibitem{Cramariuc_2023maplab2}
\BIBentryALTinterwordspacing
A.~Cramariuc, L.~Bernreiter, F.~Tschopp, M.~Fehr, V.~Reijgwart, J.~Nieto, R.~Siegwart, and C.~Cadena, ``maplab 2.0 – a modular and multi-modal mapping framework,'' \emph{IEEE Robotics and Automation Letters}, vol.~8, no.~2, p. 520–527, Feb. 2023. [Online]. Available: \url{http://dx.doi.org/10.1109/LRA.2022.3227865}
\BIBentrySTDinterwordspacing

\bibitem{hu2024metric3d}
M.~Hu, W.~Yin, C.~Zhang, Z.~Cai, X.~Long, H.~Chen, K.~Wang, G.~Yu, C.~Shen, and S.~Shen, ``Metric3d v2: A versatile monocular geometric foundation model for zero-shot metric depth and surface normal estimation,'' \emph{arXiv preprint arXiv:2404.15506}, 2024.

\bibitem{mast3r_arxiv24}
V.~Leroy, Y.~Cabon, and J.~Revaud, ``Grounding image matching in 3d with mast3r,'' 2024.

\bibitem{Geiger2012CVPRkitti}
A.~Geiger, P.~Lenz, and R.~Urtasun, ``Are we ready for autonomous driving? the kitti vision benchmark suite,'' in \emph{Conference on Computer Vision and Pattern Recognition (CVPR)}, 2012.

\bibitem{Burri25012016euroc}
\BIBentryALTinterwordspacing
M.~Burri, J.~Nikolic, P.~Gohl, T.~Schneider, J.~Rehder, S.~Omari, M.~W. Achtelik, and R.~Siegwart, ``The euroc micro aerial vehicle datasets,'' \emph{The International Journal of Robotics Research}, 2016. [Online]. Available: \url{http://ijr.sagepub.com/content/early/2016/01/21/0278364915620033.abstract}
\BIBentrySTDinterwordspacing

\bibitem{leutenegger2013keyframe}
S.~Leutenegger, P.~Furgale, V.~Rabaud, M.~Chli, K.~Konolige, and R.~Siegwart, ``Keyframe-based visual-inertial slam using nonlinear optimization,'' \emph{Proceedings of Robotis Science and Systems (RSS) 2013}, 2013.

\bibitem{newcombe2011dtam}
R.~A. Newcombe, S.~J. Lovegrove, and A.~J. Davison, ``Dtam: Dense tracking and mapping in real-time,'' in \emph{2011 international conference on computer vision}.\hskip 1em plus 0.5em minus 0.4em\relax IEEE, 2011, pp. 2320--2327.

\bibitem{Zhu2023NICER}
Z.~Zhu, S.~Peng, V.~Larsson, Z.~Cui, M.~R. Oswald, A.~Geiger, and M.~Pollefeys, ``Nicer-slam: Neural implicit scene encoding for rgb slam,'' in \emph{International Conference on 3D Vision (3DV)}, March 2024.

\bibitem{sarlin20superglue}
\BIBentryALTinterwordspacing
P.-E. Sarlin, D.~DeTone, T.~Malisiewicz, and A.~Rabinovich, ``{SuperGlue}: Learning feature matching with graph neural networks,'' in \emph{CVPR}, 2020. [Online]. Available: \url{https://arxiv.org/abs/1911.11763}
\BIBentrySTDinterwordspacing

\bibitem{sun2021loftr}
J.~Sun, Z.~Shen, Y.~Wang, H.~Bao, and X.~Zhou, ``Loftr: Detector-free local feature matching with transformers,'' in \emph{Proceedings of the IEEE/CVF conference on computer vision and pattern recognition}, 2021, pp. 8922--8931.

\bibitem{keetha2023anyloc}
N.~Keetha, A.~Mishra, J.~Karhade, K.~M. Jatavallabhula, S.~Scherer, M.~Krishna, and S.~Garg, ``Anyloc: Towards universal visual place recognition,'' \emph{IEEE Robotics and Automation Letters}, vol.~9, no.~2, pp. 1286--1293, 2023.

\bibitem{sarlin2019coarse}
P.-E. Sarlin, C.~Cadena, R.~Siegwart, and M.~Dymczyk, ``From coarse to fine: Robust hierarchical localization at large scale,'' in \emph{CVPR}, 2019.

\bibitem{barroso2024mickey}
A.~Barroso-Laguna, S.~Munukutla, V.~Prisacariu, and E.~Brachmann, ``Matching 2d images in 3d: Metric relative pose from metric correspondences,'' in \emph{CVPR}, 2024.

\bibitem{puigjaner2024augmentedrealitybordersachieving}
\BIBentryALTinterwordspacing
A.~G. Puigjaner, I.~Aloise, and P.~Schmuck, ``Augmented reality without borders: Achieving precise localization without maps,'' 2024. [Online]. Available: \url{https://arxiv.org/abs/2408.17373}
\BIBentrySTDinterwordspacing

\bibitem{duisterhof2024mast3r}
B.~Duisterhof, L.~Zust, P.~Weinzaepfel, V.~Leroy, Y.~Cabon, and J.~Revaud, ``Mast3r-sfm: a fully-integrated solution for unconstrained structure-from-motion,'' \emph{arXiv preprint arXiv:2409.19152}, 2024.

\bibitem{murai2024_mast3rslam}
R.~Murai, E.~Dexheimer, and A.~J. Davison, ``{MASt3R-SLAM}: Real-time dense {SLAM} with {3D} reconstruction priors,'' \emph{arXiv preprint}, 2024.

\bibitem{GalvezTRO12DBoW2}
D.~G\'alvez-L\'opez and J.~D. Tard\'os, ``Bags of binary words for fast place recognition in image sequences,'' \emph{IEEE Transactions on Robotics}, vol.~28, no.~5, pp. 1188--1197, October 2012.

\bibitem{cieslewski2018data}
T.~Cieslewski, S.~Choudhary, and D.~Scaramuzza, ``Data-efficient decentralized visual slam,'' in \emph{2018 IEEE international conference on robotics and automation (ICRA)}.\hskip 1em plus 0.5em minus 0.4em\relax IEEE, 2018, pp. 2466--2473.

\bibitem{Ali-bey_2024_CVPR}
A.~Ali-bey, B.~Chaib-draa, and P.~Gigu\`ere, ``{BoQ}: A place is worth a bag of learnable queries,'' in \emph{Proceedings of the IEEE/CVF Conference on Computer Vision and Pattern Recognition (CVPR)}, June 2024, pp. 17\,794--17\,803.

\bibitem{lindenberger2023lightglue}
P.~Lindenberger, P.-E. Sarlin, and M.~Pollefeys, ``{LightGlue: Local Feature Matching at Light Speed},'' in \emph{ICCV}, 2023.

\bibitem{ila2009information}
V.~Ila, J.~M. Porta, and J.~Andrade-Cetto, ``Information-based compact pose slam,'' \emph{IEEE Transactions on Robotics}, vol.~26, no.~1, pp. 78--93, 2009.

\bibitem{tyszkiewicz2020disk}
M.~Tyszkiewicz, P.~Fua, and E.~Trulls, ``Disk: Learning local features with policy gradient,'' \emph{Advances in Neural Information Processing Systems}, vol.~33, 2020.

\bibitem{Hartley:2003:MVG:861369}
R.~Hartley and A.~Zisserman, \emph{Multiple View Geometry in Computer Vision}, 2nd~ed.\hskip 1em plus 0.5em minus 0.4em\relax New York, NY, USA: Cambridge University Press, 2003.

\bibitem{boler23essentialPoseSLAM}
M.~Boler and S.~Martin, ``Essential poseslam: An efficient landmark-free approach to visual-inertial navigation,'' in \emph{2023 IEEE/ION Position, Location and Navigation Symposium (PLANS)}, 2023, pp. 1341--1349.

\bibitem{sola2018microlie}
\BIBentryALTinterwordspacing
J.~Sol{\`{a}}, J.~Deray, and D.~Atchuthan, ``A micro lie theory for state estimation in robotics,'' \emph{CoRR}, vol. abs/1812.01537, 2018. [Online]. Available: \url{http://arxiv.org/abs/1812.01537}
\BIBentrySTDinterwordspacing

\bibitem{mapanything}
N.~Keetha, N.~M\"uller, J.~Sch\"onberger, L.~Porzi, Y.~Zhang, T.~Fischer, A.~Knapitsch, D.~Zauss, E.~Weber, N.~Antunes, J.~Luiten, M.~Lopez-Antequera, S.~R. Bul\`o, C.~Richardt, D.~Ramanan, S.~Scherer, and P.~Kontschieder, ``{MapAnything}: Universal feed-forward metric {3D} reconstruction,'' in \emph{arXiv:2509.13414}, 2025.

\bibitem{abate2024kimera2robustaccuratemetricsemantic}
\BIBentryALTinterwordspacing
M.~Abate, Y.~Chang, N.~Hughes, and L.~Carlone, ``Kimera2: Robust and accurate metric-semantic slam in the real world,'' 2024. [Online]. Available: \url{https://arxiv.org/abs/2401.06323}
\BIBentrySTDinterwordspacing

\bibitem{schoenberger2016sfm}
J.~L. Sch\"{o}nberger and J.-M. Frahm, ``Structure-from-motion revisited,'' in \emph{Conference on Computer Vision and Pattern Recognition (CVPR)}, 2016.

\bibitem{lee2013robust}
G.~H. Lee, F.~Fraundorfer, and M.~Pollefeys, ``Robust pose-graph loop-closures with expectation-maximization,'' in \emph{2013 IEEE/RSJ International Conference on Intelligent Robots and Systems}.\hskip 1em plus 0.5em minus 0.4em\relax IEEE, 2013, pp. 556--563.

\bibitem{rosinol2020kimera}
A.~Rosinol, M.~Abate, Y.~Chang, and L.~Carlone, ``Kimera: an open-source library for real-time metric-semantic localization and mapping,'' in \emph{2020 IEEE International Conference on Robotics and Automation (ICRA)}.\hskip 1em plus 0.5em minus 0.4em\relax IEEE, 2020, pp. 1689--1696.

\bibitem{schubert2018vidataset}
D.~Schubert, T.~Goll, N.~Demmel, V.~Usenko, J.~Stueckler, and D.~Cremers, ``The tum vi benchmark for evaluating visual-inertial odometry,'' in \emph{International Conference on Intelligent Robots and Systems (IROS)}, October 2018.

\bibitem{brizi2024vbrvisionbenchmarkrome}
\BIBentryALTinterwordspacing
L.~Brizi, E.~Giacomini, L.~D. Giammarino, S.~Ferrari, O.~Salem, L.~D. Rebotti, and G.~Grisetti, ``Vbr: A vision benchmark in rome,'' 2024. [Online]. Available: \url{https://arxiv.org/abs/2404.11322}
\BIBentrySTDinterwordspacing

\bibitem{qin2019bvinsfusion}
T.~Qin, S.~Cao, J.~Pan, and S.~Shen, ``A general optimization-based framework for global pose estimation with multiple sensors,'' 2019.

\bibitem{Leutenegger2011brisk}
S.~Leutenegger, M.~Chli, and R.~Y. Siegwart, ``Brisk: Binary robust invariant scalable keypoints,'' in \emph{2011 International Conference on Computer Vision}, 2011, pp. 2548--2555.

\bibitem{detone2017superpoint}
\BIBentryALTinterwordspacing
D.~DeTone, T.~Malisiewicz, and A.~Rabinovich, ``Superpoint: Self-supervised interest point detection and description,'' \emph{CoRR}, vol. abs/1712.07629, 2017. [Online]. Available: \url{http://arxiv.org/abs/1712.07629}
\BIBentrySTDinterwordspacing

\bibitem{sarlin2022lamar}
P.-E. Sarlin, M.~Dusmanu, J.~L. Schönberger, P.~Speciale, L.~Gruber, V.~Larsson, O.~Miksik, and M.~Pollefeys, ``{LaMAR: Benchmarking Localization and Mapping for Augmented Reality},'' in \emph{ECCV}, 2022.

\end{thebibliography}


\end{document}